\documentclass[sigconf]{acmart}

\usepackage{booktabs}       
\usepackage{tabularx}       
\usepackage{multirow}       
\usepackage{array}          
\usepackage{multicol}
\usepackage{siunitx}  
\usepackage{enumitem}
\usepackage{bm, booktabs, amsmath}
\usepackage{pifont}

\usepackage[table]{xcolor}
\usepackage[capitalize]{cleveref}
\crefname{section}{Sec.}{Secs.}
\Crefname{section}{Section}{Sections}
\Crefname{table}{Table}{Tables}
\crefname{table}{Tab.}{Tabs.}
\usepackage{makecell}
\usepackage{booktabs}
\usepackage{tabularx}
\usepackage{array}
\usepackage{xcolor}
\usepackage{pifont}
\usepackage{tikz}
\usepackage{algorithm}
\usepackage{algpseudocode}

\newcommand{\cmark}{\textcolor{green!60!black}{\ding{51}}}
\newcommand{\xmark}{\textcolor{red!70!black}{\ding{55}}}

\definecolor{gold}{RGB}{212,175,55}
\definecolor{silver}{RGB}{160,160,160}
\definecolor{bronze}{RGB}{176,114,25}

\newcolumntype{Y}{>{\centering\arraybackslash}X}

\newcommand{\cmarkpure}{\ding{51}} 
\newcommand{\xmarkpure}{\ding{55}} 

\definecolor{good}{HTML}{4BA8C6} 
\definecolor{mid}{HTML}{8E80C6}  
\definecolor{bad}{HTML}{A7B0B8}  
\DeclareRobustCommand{\highsym}{%
  \tikz[baseline=-0.6ex]\fill[good] (0,0) circle (0.9ex);}

\DeclareRobustCommand{\medsym}{%
  \tikz[baseline=-0.6ex]{
    \draw[mid, line width=0.5pt] (0,0) circle (0.9ex);
    \fill[mid] (0,0) -- (90:0.9ex)
      arc[start angle=90,end angle=270,radius=0.9ex] -- cycle;
  }}

\DeclareRobustCommand{\lowsym}{%
  \tikz[baseline=-0.6ex]\draw[bad, line width=0.5pt] (0,0) circle (0.9ex);}

\definecolor{rayzergreen}{RGB}{88,160,120}
\definecolor{erayzerblue}{RGB}{82,140,214}
\definecolor{irisorange}{RGB}{230,150,60}
\definecolor{headerbg}{RGB}{245,247,250}

\newcommand{\equalmark}{\textsuperscript{\textdagger}}
\newcommand{\corrmark}{\textsuperscript{*}}

\newcommand{\firstpagenote}{%
  \begingroup
  \renewcommand\thefootnote{}%
  \footnotetext{%
    \equalmark\ Equal contribution.
    \corrmark\ Corresponding authors.
  }%
  \endgroup
}

\AtBeginDocument{%
  }

\copyrightyear{2026}
\acmYear{2026}
\setcopyright{cc}
\setcctype{by}
\acmConference[MM '26]{Proceedings of the 34th ACM International Conference on Multimedia}{November 10--14, 2026}{Rio de Janeiro, Brazil}
\acmBooktitle{Proceedings of the 34th ACM International Conference on Multimedia (MM '26), November 10--14, 2026, Rio de Janeiro, Brazil}
\acmDOI{10.1145/3767308.3835537}
\acmISBN{979-8-4007-2213-4/2026/11}

\acmSubmissionID{3517}

\usepackage{xcolor}
\hypersetup{
  colorlinks=true,
  urlcolor=blue,
  linkcolor=blue,
  citecolor=blue
}

\begin{document}

\title{IRIS: Implicit Rendering Matters for Pose-Free Novel View Synthesis}


\author{Wenyu Li\equalmark, Sidun Liu\equalmark, Peng Qiao\corrmark, Yong Dou\corrmark, and Tongrui Hu}
\affiliation{%
  \institution{National University of Defense Technology}
  \city{Changsha}
  \country{China}}


\begin{abstract}
Novel view synthesis from unposed multi-view images remains challenging, as the model must jointly learn scene representations and camera parameters without pose supervision. 
Existing approaches largely fall into two extremes: implicit latent-space rendering is flexible and easy to optimize, but often yields weakly grounded camera estimation; explicit 3D representations provide stronger geometric grounding, but introduce heavier parameterization and more fragile optimization. 
In this paper, we present IRIS, a fully self-supervised framework that provides a practical middle ground between these two paradigms. 
Instead of decoding free latent tokens or reconstructing fully explicit 3D primitives, IRIS represents the scene as a latent neural field and renders novel views by querying this field under self-predicted cameras. 
Specifically, projected features from reference views are aggregated at sampled 3D points to form point-wise latent features, which are then composed along target rays for rendering. 
This design preserves the flexibility and optimization stability of implicit modeling, while introducing stronger geometric structure than unconstrained latent rendering. 
Extensive experiments show that IRIS achieves strong novel view synthesis quality with competitive pose accuracy under fully self-supervised learning.
Our project page:
\href{https://leo-frank.github.io/IRIS/}{\textcolor{blue}{\url{https://leo-frank.github.io/IRIS/}}}.
\end{abstract}





\maketitle
\firstpagenote

\section{Introduction}
Novel view synthesis is a fundamental problem in computer vision and computer graphics, aiming to recover scene representation from limited observations and support high-quality rendering at target viewpoints. In recent years, this area has witnessed remarkable progress, ranging from radiance field representations such as NeRF\cite{mildenhall20nerf:}, to explicit scene representations such as 3D Gaussian Splatting\cite{kerbl3Dgaussians}, and further to feed-forward models\cite{srt22,yu21pixelnerf:,jin24lvsm:,jiang2025anysplat,GNT,gntmove2023} for view synthesis without scene-specific optimization.

Most of these methods rely on known or pre-estimated camera poses\cite{GNT,gntmove2023,chen21mvsnerf:,jin24lvsm:}, which directly affect the quality of target-view rendering. Their dependence on external camera acquisition still poses clear limitations. In practice, camera parameters are often obtained through Structure-from-Motion pipelines\cite{schonberger16structure-from-motion,schonberger16pixelwise}, which not only increase computational cost and engineering complexity, but also become unstable in the presence of weak textures, heavy occlusions, or large viewpoint changes. Moreover, such dependence on external pose estimation limits the scalability of these methods to large collections of raw multi-view data. As a result, directly learning from unposed images, namely pose-free or self-calibrated novel view synthesis\cite{sajjadi2022rust, jiang2025rayzer}, is becoming increasingly important.

Existing pose-free methods\cite{jiang2025rayzer,zhao2026erayzer,huang2025spfsplat} have shown that it is feasible to jointly learn scene representations and camera parameters without pose supervision, but their rendering designs still largely fall into two extremes.
The pioneering RayZer~\cite{jiang2025rayzer} adopts a flexible \textit{implicit} latent-space renderer and is easier to optimize. However, its predicted cameras are more likely to degenerate into latent variables that merely support internal rendering compatibility, rather than corresponding to physically grounded geometry.
Its follow-up, E-RayZer~\cite{zhao2026erayzer}, obtains more physically-grounded camera spaces by representing the scene with \textit{explicit} 3D Gaussians, but at the cost of lower rendering quality and more sensitive optimization.
This raises a natural question: can one simultaneously retain the advantages of both sides, namely high rendering quality and more physically-grounded camera estimation?

Our key insight is that geometric constraints remain essential for self-supervised novel view synthesis. However, such constraints need not be introduced only through fully explicit 3D primitives.
Instead, they can also arise from how an implicit scene representation is structured and rendered under self-predicted cameras.
Based on this observation, we propose IRIS, which represents the scene as a latent neural field and renders novel views by querying this field under self-predicted cameras.
Concretely, rather than decoding free latent tokens as in RayZer or reconstructing fully explicit 3D primitives as in E-RayZer, IRIS builds a latent neural field as the scene representation. 
Novel views are rendered by querying this field along target rays: at each sampled 3D point, projected features from reference views are aggregated into a point-wise latent feature, and these point-wise features are further composed along the ray to predict the target color. 
This design preserves the flexibility and trainability of implicit modeling, while introducing stronger geometric structure than unconstrained latent rendering.

Experiments show that IRIS achieves strong novel view synthesis quality while maintaining competitive pose accuracy under the fully self-supervised setting. Compared with prior methods, IRIS is more robust to unordered inputs while avoiding the optimization difficulty of fully explicit 3D representations. Our main contributions can be summarized as follows: (1) We propose IRIS, a fully self-supervised framework for novel view synthesis from unposed multi-view images, based on a latent neural field representation and field querying under self-predicted cameras. (2) We present a field-based alternative to existing self-supervised rendering designs, which avoids both unconstrained latent decoding and fully explicit 3D primitives. (3) We show through experiments that IRIS achieves strong novel view synthesis quality and pose accuracy.

\begin{table}[t]
\centering
\small
\renewcommand{\arraystretch}{1.15}
\setlength{\tabcolsep}{9pt}
\begin{tabular}{lccc}
\toprule
Property & RayZer\cite{jiang2025rayzer} & E-RayZer\cite{zhao2026erayzer} & IRIS \\
\midrule
Ordered-input NVS   & \highsym & \medsym  & \highsym \\
Unordered-input NVS & \lowsym  & \medsym  & \highsym \\
Grounded camera    & \lowsym  & \highsym & \highsym \\
Trainability       & \highsym & \lowsym  & \highsym \\
\bottomrule
\end{tabular}
\caption{High-level comparison of RayZer, E-RayZer, and IRIS.
\texorpdfstring{\highsym}{strong},
\texorpdfstring{\medsym}{moderate}, and
\texorpdfstring{\lowsym}{weak}
denote strong, moderate, and weak performance, respectively.}
\label{tab:simple_compare}
\vspace{-2em}
\end{table}


\begin{figure*}[t]
    \centering
    \includegraphics[width=\textwidth]{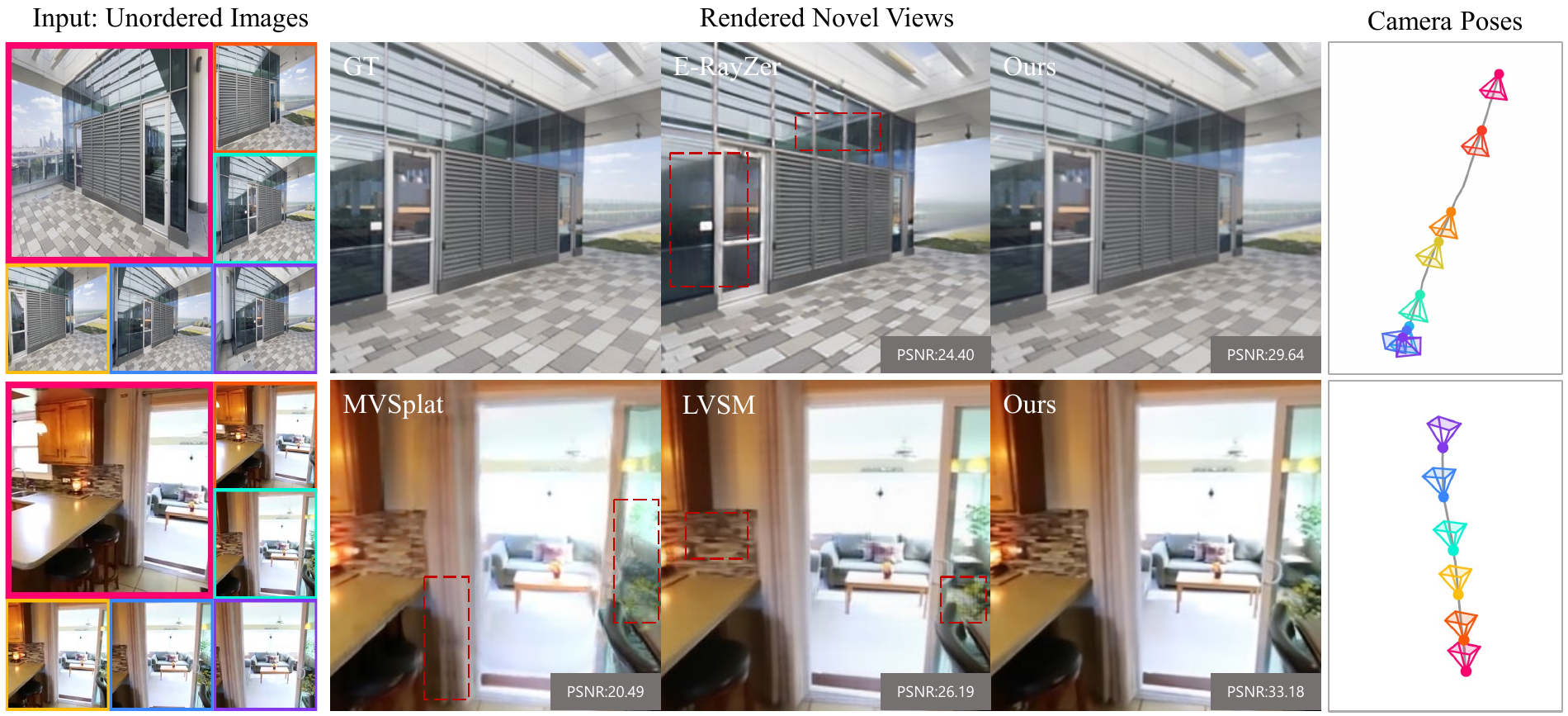}
    \caption{
   Given a set of unordered input images, our method predicts camera poses and synthesizes high-quality novel views, while being trained entirely on images without camera annotations. Compared with prior methods, our approach produces sharper and more faithful renderings. The rightmost panels visualize the predicted camera poses.}
    \label{fig:teaser}
\end{figure*}

\section{Related Work}
In this section, we briefly review prior work on novel view synthesis (NVS). From the perspective of pose dependency, existing methods can be broadly categorized into three groups:
\textbf{(1) Pose-Conditioned NVS.} These methods assume camera poses are available as input and synthesize novel views from posed images.
\textbf{(2) Pose-Supervised NVS.} These methods do not require poses at inference time, but still rely on pose annotations or pose-related supervision during training.
\textbf{(3) Fully Self-Supervised NVS.} These methods learn both camera pose estimation and novel view synthesis directly from unlabeled images, without using pose annotations in either training or inference.

\textbf{Pose-Conditioned NVS.}
These methods assume camera poses are available at inference time and synthesize novel views from posed images. Representative examples include Neural Radiance Fields (NeRF~\cite{mildenhall2021nerf}) and 3D Gaussian Splatting (3DGS~\cite{kerbl3Dgaussians})
However, both original methods require costly per-scene optimization, limiting practicality. To address this issue, later works developed generalizable pose-conditioned frameworks that amortize reconstruction and
rendering across scenes. These methods can be broadly grouped into three categories
(1) \textbf{Radiance-field-based methods}~\cite{hong2023lrm,yu21pixelnerf:,wang21ibrnet:,chen21mvsnerf:,GNT,gntmove2023} query radiance or latent properties at arbitrary 3D locations by combining spatial coordinates with image features from multiple source views, and perform ray-based rendering for target-view synthesis. 
(2) \textbf{3DGS-based methods}~\cite{gslrm2024,charatan23pixelsplat:,jiang2025anysplat,ziwen2025llrm,chen24mvsplat:,chen24mvsplat360:} directly predict 3D Gaussian primitives as the scene representation and render novel views through differentiable splatting. 
(3) \textbf{Neural-network-only methods}~\cite{sajjadi21scenesrt,jin2025lvsm} remove explicit 3D inductive biases and instead learn a latent rendering function directly from data. 
Although these methods substantially reduce optimization cost and improve efficiency, their applicability is still limited by the requirement for accurate camera poses at test time. This limitation motivates later work that further relaxes pose dependency during training and inference.

\textbf{Pose-Supervised NVS.}
A second line of research relaxes the reliance on camera poses at inference time, while still using pose supervision during training. 
Compared with pose-conditioned methods, these methods learn scene representations directly from 2D images instead of taking camera poses as input. 
Trained on large-scale posed image collections, such methods have shown promising performance in novel view synthesis. 
For example, LEAP~\cite{jiang2022LEAP} and PF-LRM~\cite{wang2024pf} construct neural volumes in the canonical view coordinate system and perform neural rendering from the inferred representation. NoPoSplat~\cite{ye2024noposplat} reconstructs scene Gaussians in the coordinate system of the first view directly from input images without requiring camera poses as input.
Nevertheless, these methods still rely on ground-truth poses during training, which limits their scalability to fully unposed image collections.

\textbf{Self-Supervised NVS.}
Fully self-supervised NVS aims to eliminate the reliance on camera poses during both training and inference, and instead nstead learns camera estimation and novel view synthesis.
In practice, camera poses are often estimated using Structure-from-Motion (SfM) pipelines such as COLMAP~\cite{schops2017multi}, but these methods can be brittle under sparse-view or wide-baseline settings. Even when many views are available, incremental SfM may still discard difficult image regions by filtering feature correspondences that are inconsistent with the current reconstruction. 
With the differentiability of rendering methods, it is possible to optimize camera parameters jointly with scene representations. 
Early works such as NeRF--~\cite{wang2021nerfmm} show that camera poses can be estimated for forward-facing scenes by jointly optimizing poses and radiance fields from simple initialization, while BARF~\cite{lin2021barf} improves optimization stability through a coarse-to-fine registration strategy. 
More recently, RayZer~\cite{jiang2025rayzer} removes explicit 3D inductive bias and adopts latent rendering, achieving strong synthesis quality for ordered image sequences. However, its learned pose space is only weakly grounded in physically interpretable 3D geometry, which limits its 3D awareness. E-RayZer~\cite{zhao2026erayzer} moves one step further by introducing explicit 3D Gaussian reconstruction into the self-supervised setting. 
While such explicit 3D representations offer stronger geometric grounding, they also make training substantially more difficult.
To stabilize training, E-RayZer further adopts a fine-grained visual-overlap-based curriculum that starts from high-overlap samples and gradually expands to harder ones, while adaptively aligning heterogeneous data sources.
Our method is also situated in the fully self-supervised setting, but differs from prior approaches by seeking a better balance between geometric grounding, rendering quality, and optimization stability: instead of relying on unconstrained latent rendering or explicit 3D Gaussian reconstruction alone, we learn a latent neural field to represent the scene.

\section{Approach}

In the fully self-supervised setting, our goal is to achieve strong novel view synthesis while learning camera parameters and scene representations directly from unlabeled multi-view images.
We first revisit the two representative self-calibrated paradigms, namely RayZer\cite{jiang2025rayzer} and E-RayZer\cite{zhao2026erayzer}, and clarify their key differences. We then present \textbf{IRIS}, which combines camera prediction with a latent field representation.

\subsection{Preliminaries: Implicit or Explicit}
Given a set of unposed multi-view images $\mathcal{I} = \{ I_i \}_{i=1}^{V}$,
following prior self-supervised learning frameworks, we split the input into two non-overlapping subsets: a reference set $\mathcal{I}_{\mathrm{ref}}$ used for scene inference, and a target set $\mathcal{I}_{\mathrm{tgt}}$ used for photometric supervision. Let $\mathcal V_{\mathrm{ref}}$ and $\mathcal V_{\mathrm{tgt}}$ denote the index sets of the reference and target views, respectively, with $V_{\mathrm{ref}}=|\mathcal V_{\mathrm{ref}}|$.

\textbf{RayZer} first patchifies all images into image tokens \(f\), and introduces one learnable
camera token per view, denoted by \(p\in\mathbb R^{V\times d}\).
A transformer-based camera estimator updates them jointly:
\[
\{f^{*},p^{*}\}=E_{\mathrm{cam}}(\{f,p\}).
\]
A canonical reference view \(c\) is selected, and the relative pose of each view \(i\) with respect to
\(c\) is predicted as
\[
\pi_i=\mathrm{MLP}_{\mathrm{pose}}([p_i^{*},p_c^{*}])\in\mathbb R^9,
\qquad
P_i=\Phi_{\mathrm{SE(3)}}(\pi_i),
\]
where \(\pi_i\) consists of a 6D rotation parameterization and a 3D translation and
\(P_i\in SE(3)\) denotes the camera pose. RayZer further
predicts a shared focal length from the canonical camera token,
\[
focal=\mathrm{MLP}_{\mathrm{focal}}(p_c^{*}),
\]
which defines a shared intrinsic matrix \(K\) for all views, yielding the predicted camera parameters
\[
\mathcal P=\{(P_i,K)\}_{i=1}^V.
\]

Each predicted camera is converted into a pixel-aligned Pl\"ucker ray map
\[
R_i^{\mathrm{plk}}=\Pi_{\mathrm{plk}}(P_i,K).
\]
For the reference-view subset \(\mathcal I_{\mathrm{ref}}\), RayZer linearly tokenizes the predicted
Pl\"ucker ray maps into ray tokens \(r_{\mathrm{ref}}\), fuses them with the corresponding image tokens
\(f_{\mathrm{ref}}\),
\[
x_{\mathrm{ref}}=\mathrm{MLP}_{\mathrm{fuse}}([f_{\mathrm{ref}},r_{\mathrm{ref}}]),
\]
and predicts the latent scene representation from learnable scene tokens \(z\) via
\[
\{z^{*},x_{\mathrm{ref}}^{*}\}=E_{\mathrm{scene}}(\{z,x_{\mathrm{ref}}\}),
\qquad
z_{\mathrm{scene}}=z^{*}.
\]
Finally, the target views are rendered from the latent scene representation and the target-view ray
tokens,
\[
\hat I_{\mathrm{tgt}}=f_{\phi}^{\mathrm{rend}}(z_{\mathrm{scene}},r_{\mathrm{tgt}}),
\]
and the model is trained with photometric supervision on target views:
\[
\mathcal L=\sum_{(I,\hat I)\in(\mathcal I_{\mathrm{tgt}},\hat{\mathcal I}_{\mathrm{tgt}})}
\Bigl(
\mathrm{MSE}(\hat I,I)+\lambda_{\mathrm{perc}}\,\mathrm{Percep}(\hat I,I)
\Bigr).
\]

\textbf{E-RayZer} follows the same overall pose-first formulation as RayZer, but replaces
RayZer’s implicit latent scene representation with \emph{explicit} 3D Gaussians~\cite{kerbl3Dgaussians}.
Using the same fused image--ray tokens \(x_{\mathrm{ref}}\) as in RayZer, E-RayZer predicts the Gaussian
scene as
\[
\mathcal{G}_{\mathrm{ref}}
=
\bigl(f_{\omega}^{\mathrm{gauss}} \circ E_{\mathrm{scene}}\bigr)(x_{\mathrm{ref}})
\]
Here, \(E_{\mathrm{scene}}(x_{\mathrm{ref}})\) denotes the scene encoding stage that performs multi-view
aggregation and produces updated latent tokens, while \(f_{\omega}^{\mathrm{gauss}}(\cdot)\) is a lightweight
Gaussian decoder that maps these tokens to per-pixel Gaussian parameters $\{g_i=(d_i,q_i,C_i,s_i,\alpha_i)\}_{i=1}^{V_{\mathrm{ref}}HW}$ along the reference-view rays.
Each Gaussian \(g_i\) is parameterized by distance along the ray \(d_i\), orientation \(q_i\), spherical-harmonic
color coefficients \(C_i\), scale \(s_i\), and opacity \(\alpha_i\). The target views are then rendered using the
predicted target-view cameras
\[
P_{\mathrm{tgt}}=\{(K,P_i)\mid i\in \mathcal V_{\mathrm{tgt}}\},
\qquad
\hat I_{\mathrm{tgt}}=\pi(\mathcal G_{\mathrm{ref}}, P_{\mathrm{tgt}})
\]
where \(\pi(\cdot)\) denotes differentiable Gaussian splatting.

\begin{figure*}[t]
    \centering
    \includegraphics[width=0.7\textwidth]{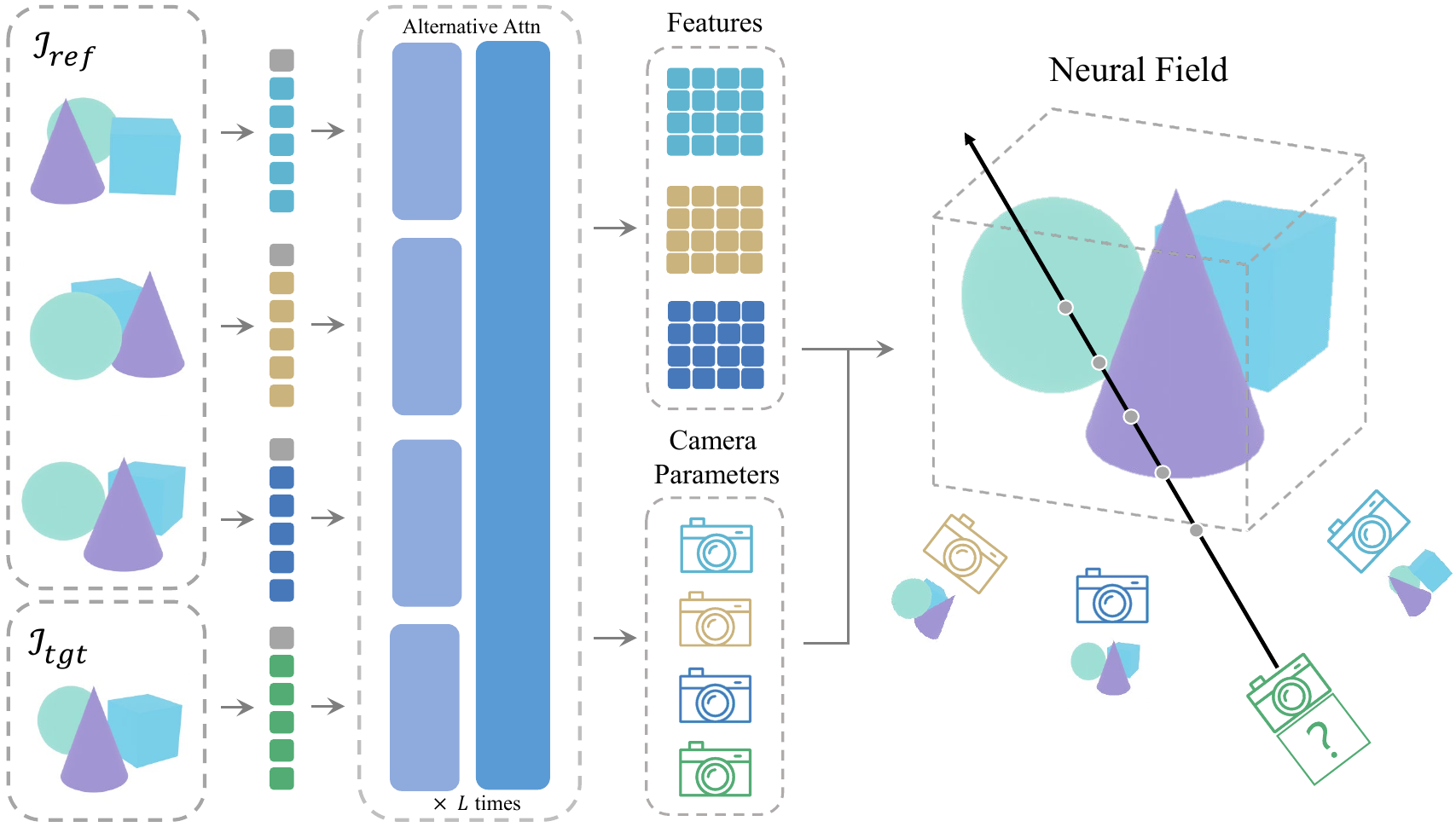}
    \caption{Overview of our method. 
    Given unposed multi-view images, IRIS first predicts camera parameters and extracts latent reference features with a shared multi-view encoder, from which a latent neural field is constructed.
    Novel views are rendered by querying this field along target rays under the self-predicted cameras, where point-wise features are aggregated across reference views and then composed along each ray for color prediction.
    }
    \label{fig:method}
\end{figure*}

\textbf{Comparison and Motivation.}
RayZer benefits from flexible latent-space rendering and favorable optimization, but because camera prediction, scene inference, and rendering are jointly learned in \textit{latent space}, the predicted cameras mainly need to remain \textit{mutually compatible} with the internal renderer. As a result, the learned camera space is not necessarily physically grounded, and may partially degenerate into a latent variable that mainly serves rendering compatibility. By replacing latent rendering with \textit{explicit} 3D Gaussians and differentiable splatting, E-RayZer achieves a more geometrically grounded camera space. This improvement, however, comes at the cost of a heavier representation, a more sensitive optimization process, and slightly weaker rendering quality. 
These two paradigms therefore expose a clear trade-off between optimization flexibility and geometric grounding. Our goal is to preserve the optimization advantages of latent scene modeling, while replacing unconstrained latent decoding with a rendering interface that is more strongly constrained by multi-view geometric consistency.

\subsection{IRIS Model}

\textbf{Our Insight.}
The above comparison suggests that some form of geometric inductive bias remains essential for self-supervised novel view synthesis. Without it, camera prediction and scene representation may drift toward a latent solution that is internally compatible with the renderer, but not physically grounded. At the same time, simply adopting fully explicit 3D representations is not ideal either, since it introduces a heavier representation and a more fragile optimization process. Our key idea is therefore to introduce geometric structure in a lighter-weight manner: instead of relying on unconstrained latent decoding as in RayZer, or fully explicit 3D primitives as in E-RayZer, IRIS represents the scene as a \textit{latent field}, which is queried and rendered under self-predicted cameras.

\textbf{Shared Multi-View Encoder.}
Following RayZer and E-RayZer, IRIS adopts a \emph{pose-first} formulation: camera parameters are predicted first and then used to define the scene for rendering.
Unlike RayZer, however, IRIS does not use a dedicated camera estimator followed by a separate scene
reconstructor conditioned on fused image--ray tokens. 
Instead, we use a \emph{single shared multi-view encoder} to jointly produce per-view visual features and camera tokens, and directly retain the encoded view features for the subsequent latent-field representation.
We instantiate the shared encoder with Muskie~\cite{li2025muskiemultiviewmaskedimage}, since
Muskie is a backbone designed for multi-view inputs that processes all input views jointly rather than encoding each
frame independently. 
Given the input images $\mathcal{I}=\{I_i\}_{i=1}^{V}$
we tokenize each view into patch tokens and prepend one learnable camera token \(p_i\) to each view.
The resulting multi-view tokens are processed jointly by the shared encoder:
\[
\{f_i,\; p_i^{*}\}_{i=1}^{V}
=
E_{\theta}^{\mathrm{enc}}(\mathcal{I}),
\]
where \(f_i\in\mathbb{R}^{hw\times d}\) denotes the encoded feature map of the \(i\)-th view, and
\(p_i^{*}\in\mathbb{R}^{d}\) denotes its updated camera token.
A camera head then predicts camera parameters in the same way as RayZer: a canonical reference
view \(c\) is selected, each view pose is regressed relative to \(c\) from the updated camera tokens,
and a single shared intrinsic matrix \(K\) is predicted for all views from the canonical token. We denote
the resulting camera set by $\mathcal{P}=\{(P_i,K)\}_{i=1}^{V}$.

\textbf{Scene Representation.}
Unlike RayZer and E-RayZer, IRIS does \emph{not} reconstruct a separate set of latent scene tokens
or explicit 3D primitives. 
Instead, IRIS directly preserves the reference-view features produced by the shared encoder and
uses them, together with the self-predicted cameras, as a latent scene memory. Formally, we denote
this memory by
\[
\mathcal{S}
=
\{(f_k,K,P_k)\}_{k\in\mathcal{V}_{\mathrm{ref}}}
\]

This scene memory induces a latent field, whose value at a 3D query point \(x\) is obtained by
projecting \(x\) into all reference views under the predicted cameras, sampling the corresponding
view-wise features, and aggregating them across views. Concretely, for each reference view \(k\), we
first obtain
\[
v_k(x)=\Pi_k(x;f_k,P_k,K),
\]
where \(\Pi_k(\cdot)\) denotes projection followed by feature sampling from the \(k\)-th reference-view
feature map. We then define the latent feature of \(x\) as
\[
\tilde f(x)
=
F_{\mathcal S}(x)
=
\mathcal{T}_{\mathrm{view}}\!\bigl(v_1(x),\dots,v_{V_{\mathrm{ref}}}(x)\bigr),
\]
where \(\mathcal{T}_{\mathrm{view}}\) is a transformer that fuses the projected features from all reference
views into a single point-wise latent feature. In this sense, \(F_{\mathcal S}\) is a latent field induced on the fly from multi-view features under the self-predicted cameras.

\textbf{Rendering.}
Given a target ray \(r=(o_r,d_r)\), we sample \(M\) points along the ray,
\[
x_m=o_r+z_m d_r,\qquad m=1,\dots,M,
\]
and query the latent field at each sampled point:
\[
\tilde f_m = F_{\mathcal S}(x_m).
\]
We then compose the resulting point-wise features along the ray with a second transformer,
\[
h_r
=
\mathcal{T}_{\mathrm{ray}}(\tilde f_1,\dots,\tilde f_M),
\]
where \(\mathcal{T}_{\mathrm{ray}}\) is a transformer that aggregates the sampled point features into a ray-wise feature. Unlike classical volumetric rendering, which predicts explicit color and density
for each sample and combines them with a fixed rendering rule\cite{mildenhall2021nerf}, \(\mathcal{T}_{\mathrm{ray}}\) learns how the queried latent features should be composed to form the final ray representation.
The rendered color is then predicted as
\[
\hat C(r)=f_{\mathrm{rgb}}(h_r).
\]
We train IRIS with self-supervised photometric loss on target rays:
\[
\mathcal L
=
\sum_{r\in\mathcal R_{\mathrm{tgt}}}
\|\hat C(r)-C(r)\|_2^2.
\]
Since both latent-field querying and ray-wise rendering are conditioned on the self-predicted cameras,
this objective jointly supervises camera estimation, scene representation learning, and rendering in a
fully self-supervised manner.

Overall, IRIS can be viewed as a middle ground between RayZer and E-RayZer. Unlike RayZer, it does not rely on unconstrained latent decoding for rendering; unlike E-RayZer, it avoids fully explicit 3D primitives and their associated optimization difficulty. Instead, IRIS introduces geometric structure by constraining how the latent scene representation is queried and rendered under self-predicted cameras.



\begin{figure*}[t]
    \centering
    \includegraphics[width=0.8\textwidth]{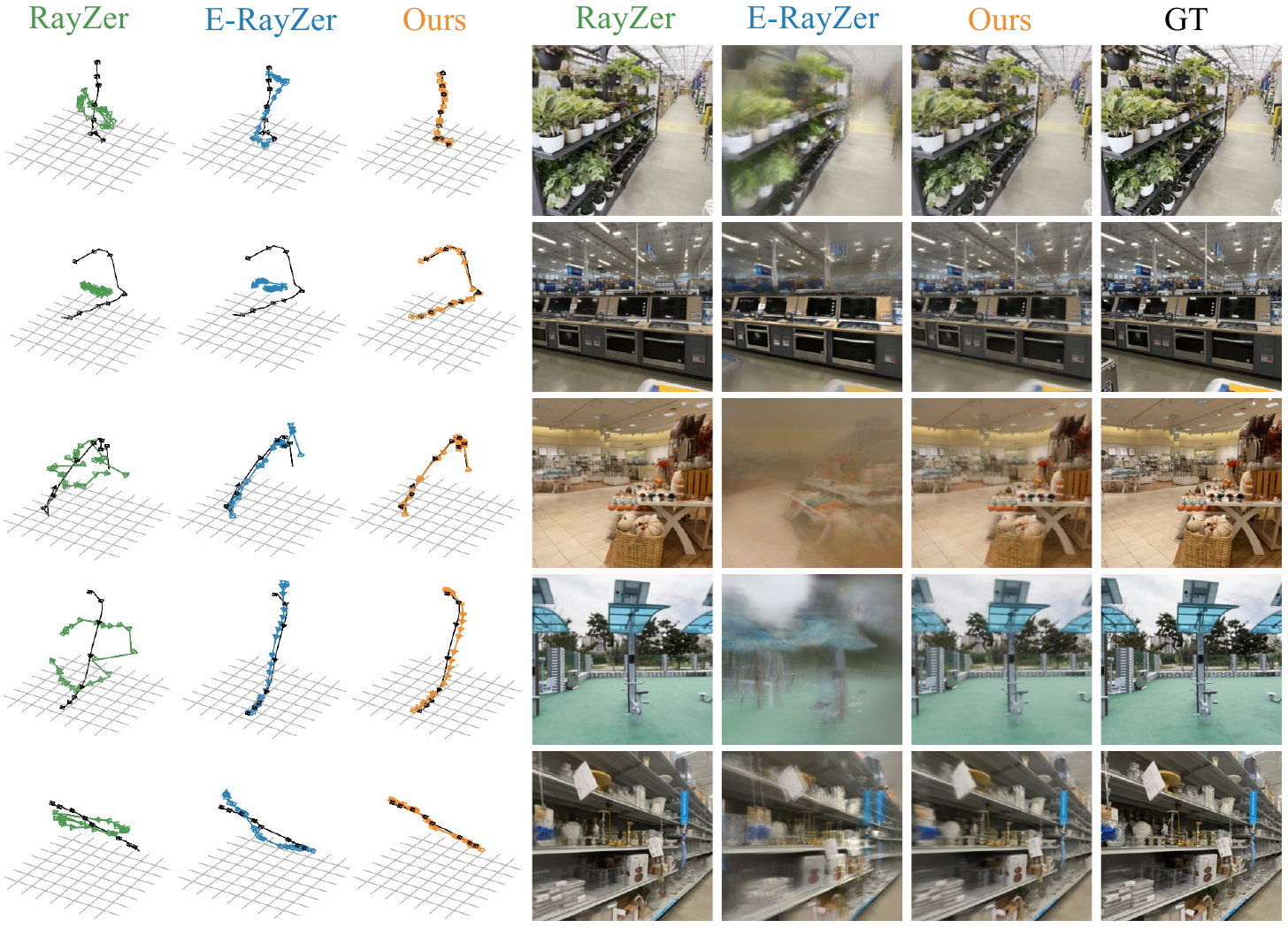}
    \caption{Qualitative comparisons of pose estimation and rendering quality. We visualize all poses of the predicted trajectory as colored frustums, while for the COLMAP trajectory we display only every third pose as a black frustum to reduce overlap.}
    \label{fig:nvs_and_pose}
\end{figure*}
\section{Experiments}

\subsection{Experimental Setup}

We report evaluation results for quality of novel view synthesis and pose estimation on several datasets to demonstrate the effectiveness of our method.

{\textbf{Implementation Details.}}
We use Muskie\cite{li2025muskiemultiviewmaskedimage} as the visual backbone and fine-tune it during training. 
For rendering, we sample 64 points within a fixed depth range along each ray. During training, we randomly sample a subset of rays for efficiency, while at inference time full-resolution images are rendered in chunks.
We optimize the model with AdamW and a base learning rate of \(4\times10^{-5}\), together with a warmup schedule. The encoder and pose predictor are trained with a \(0.1\times\) learning rate, while the renderer uses the full rate. 
Training is conducted on 8 A100 GPUs.
In practice, training on DL3DV takes roughly 7 days, while training on the mixed-dataset setting takes about 14 days.

{\textbf{Training and Testing Data.}}
We report results under both \textbf{single-dataset} and \textbf{multi-dataset} training settings. 
For single-dataset training, models are trained exclusively on Re10K~\cite{re10k} or DL3DV~\cite{ling2024dl3dv}. 
For multi-dataset training, we train on a mixture of Re10K~\cite{re10k}, DL3DV~\cite{ling2024dl3dv}, CO3Dv2~\cite{reizenstein21co3d}, ARKitScenes~\cite{dehghan2021arkitscenes}, and WildRGBD~\cite{xia2024wildrgbd}, covering a diverse set of indoor and outdoor scenes. 
This training setup is comparable to that of E-RayZer\cite{zhao2026erayzer} in terms of both dataset scale and diversity.
For evaluation, we mainly benchmark both pose estimation and novel view synthesis on the DL3DV~\cite{ling2024dl3dv} test set, NRGBD~\cite{nrgbd_dataset_cvpr22}, 7Scenes~\cite{7scenes}, and ScanNet++~\cite{yeshwanthliu2023scannetpp}.
Each test sample contains 24 frames, with 16 used as reference and the remaining 8 as target views. We evaluate pose estimation over all frames, while NVS quality is reported only on the 8 target views.
To better reflect real-world usage, our evaluation primarily uses randomly sampled input images that do not assume any fixed temporal or pose ordering; we denote this setting as \textbf{Rand}. In addition, since we find that RayZer is only applicable to ordered image sequences due to its internal image index embeddings, we also report results on ordered input sequences, denoted as \textbf{Seq}, to enable a more complete comparison with RayZer.

{\textbf{Baselines.}}
We compare with baselines from three categories: pose-conditioned methods (MVSplat~\cite{chen24mvsplat:} and LVSM~\cite{jin24lvsm:}), pose-supervised methods (NoPoSplat~\cite{ye2024noposplat}), and self-supervised methods~(SelfSplat~\cite{kang2025selfsplat}, SPFSplat~\cite{huang2025spfsplat}, RayZer~\cite{jiang2025rayzer}, and E-RayZer~\cite{zhao2026erayzer}). 
We do not include comparisons with PF-LRM\cite{wang2024pf}, since its official implementation is not publicly available.
We note that NoPoSplat and SelfSplat are primarily designed for two-view input and show limited scalability to the multi-view setting. Moreover, among the self-supervised baselines, only RayZer and E-RayZer have been trained at relatively large scale---RayZer on DL3DV, and E-RayZer on an even broader mixture of datasets---whereas other methods (e.g., SPFSplat) are mainly pretrained on Re10K~\cite{re10k}, whose camera trajectories are considerably less diverse. Therefore, we regard RayZer and E-RayZer as the primary baselines in our comparisons.

{\textbf{Evaluation Metrics.}}
We evaluate novel view synthesis using the standard metrics PSNR, SSIM, and LPIPS.
For pose estimation, we report relative pose accuracy (RPA) at thresholds of \(5^\circ\), \(15^\circ\), and \(30^\circ\), where the relative pose error is defined as the maximum of the rotation error and the translation-direction error between the predicted and ground-truth relative poses.

\begin{table}[t]
\centering
\small
\resizebox{\columnwidth}{!}{
\begin{tabular}{llcccccccc}
\toprule
\multirow{2}{*}{Dataset} & \multirow{2}{*}{Method} & \multirow{2}{*}{Self-supervised?}
& \multicolumn{3}{c}{Pose Accuracy}
& \multicolumn{3}{c}{Novel View Synthesis} \\
\cmidrule(lr){4-6} \cmidrule(lr){7-9}
& & & @5$^\circ\uparrow$ & @15$^\circ\uparrow$ & @30$^\circ\uparrow$ & PSNR$\uparrow$ & LPIPS$\downarrow$ & SSIM$\uparrow$ \\
\midrule

\multirow{2}{*}{Re10K}
& SPFSplat & \xmarkpure (MASt3R\cite{mast3r})
& 0.673 & 0.928 & 0.961 & 24.418 & 0.148 & 0.788 \\
& Ours     & \cmarkpure & 0.590 & 0.884 & 0.997 & 31.272 & 0.117 & 0.920 \\

\bottomrule
\end{tabular}
}
\caption{Comparison under the single-source training setting on Re10K\cite{re10k}.}
\label{tab:re10k_in_dataset}
\vspace{-2em}
\end{table}
\begin{table}[!t]
\centering
\small
\resizebox{\columnwidth}{!}{
\begin{tabular}{llcccccc}
\toprule
\multirow{2}{*}{Dataset} & \multirow{2}{*}{Method} 
& \multicolumn{3}{c}{Pose Accuracy} 
& \multicolumn{3}{c}{Novel View Synthesis} \\
\cmidrule(lr){3-5} \cmidrule(lr){6-8}
& & @5$^\circ\uparrow$ & @15$^\circ\uparrow$ & @30$^\circ\uparrow$ & PSNR$\uparrow$ & LPIPS$\downarrow$ & SSIM$\uparrow$ \\
\midrule

\multirow{3}{*}{\shortstack{DL3DV\cite{ling2024dl3dv}\\(Seq.)}}
& RayZer    & 0.000 & 0.010 & 0.089 & 26.286 & 0.169 & 0.814 \\
& E-RayZer  & 0.658 & 0.824 & 0.902 & 21.187 & 0.272 & 0.701 \\
& Ours      & 0.600 & 0.869 & 0.918 & 25.402 & 0.215 & 0.791 \\
\midrule

\multirow{3}{*}{\shortstack{DL3DV\cite{ling2024dl3dv}\\(Rand.)}}
& RayZer    & \multicolumn{6}{c}{\textit{Fail}} \\
& E-RayZer  & 0.656 & 0.811 & 0.895 & 21.124 & 0.271 & 0.701 \\
& Ours      & 0.563 & 0.863 & 0.924 & 25.324 & 0.215 & 0.791 \\
\midrule

\multirow{2}{*}{NRGBD\cite{nrgbd_dataset_cvpr22}}
& E-RayZer  & 0.362 & 0.800 & 0.916 & 26.476 & 0.144 & 0.860 \\
& Ours      & 0.437 & 0.861 & 0.979 & 31.404 & 0.128 & 0.921 \\
\midrule

\multirow{2}{*}{ScanNet++\cite{yeshwanthliu2023scannetpp}}
& E-RayZer  & 0.019 & 0.252 & 0.545 & 21.992 & 0.259 & 0.744 \\
& Ours      & 0.036 & 0.333 & 0.627 & 22.938 & 0.276 & 0.763 \\
\midrule

\multirow{2}{*}{7Scenes\cite{7scenes}}
& E-RayZer  & 0.312 & 0.729 & 0.882 & 26.727 & 0.175 & 0.870 \\
& Ours      & 0.278 & 0.749 & 0.856 & 30.783 & 0.140 & 0.907 \\

\bottomrule
\end{tabular}
}
\caption{Comparison under the single-source training setting on DL3DV\cite{ling2024dl3dv}. RayZer fails when testing with random-order views.}
\label{tab:train_dl3dv_comparison}
\vspace{-2em}
\end{table}

\subsection{Results}

\begin{figure}[!t]
    \centering
    \includegraphics[width=0.99\columnwidth]{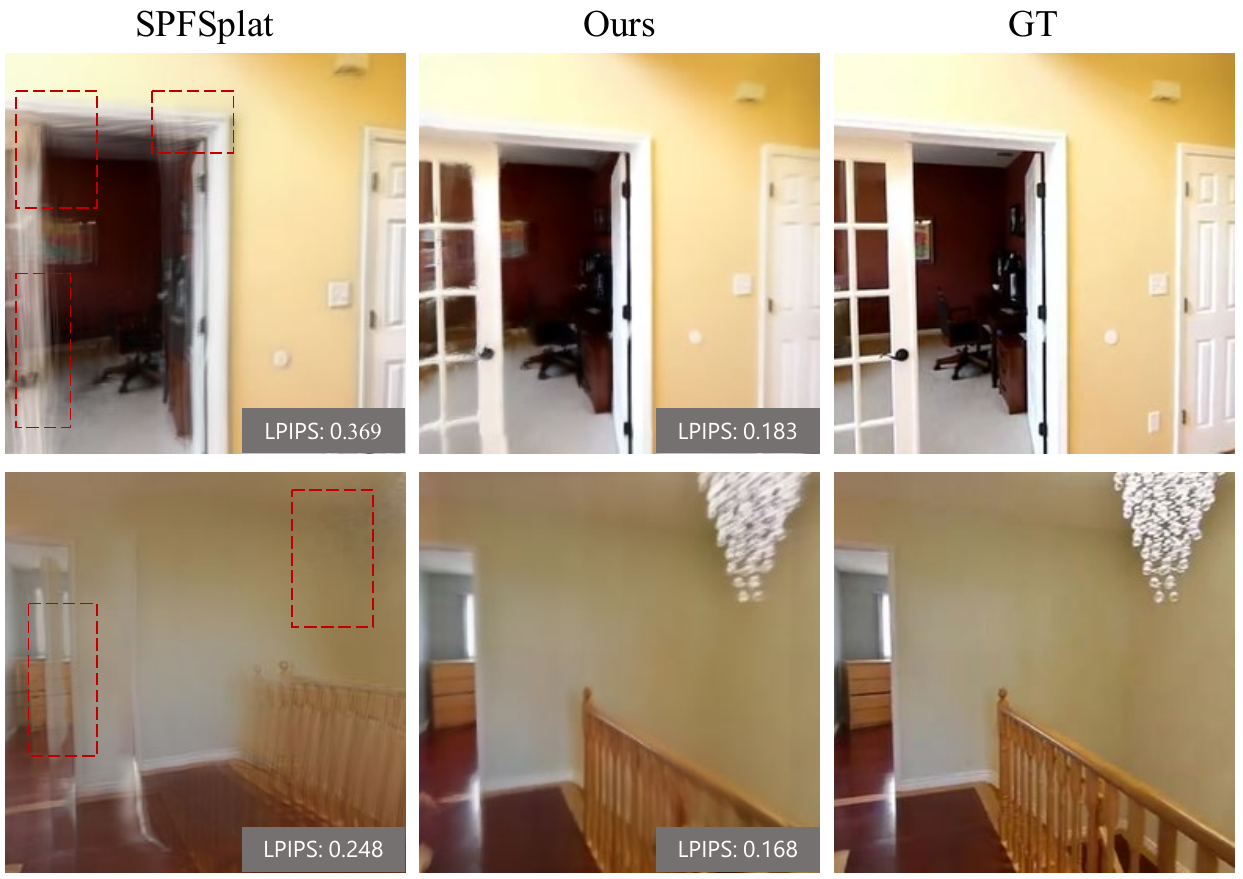}
    \caption{Qualitative comparisons with SPFSplat\cite{huang2025spfsplat} on Re10K\cite{re10k}. Our method produces sharper and more structurally consistent renderings, achieving lower LPIPS.}
    \label{fig:re10k}
\end{figure}

\textbf{Main Results.}
We first evaluate our method under the \textbf{single-dataset} training setting, where the model is trained on a single dataset.
We begin with Re10K, a widely used benchmark with relatively regular camera trajectories.
As shown in \cref{tab:re10k_in_dataset} and \cref{fig:re10k}, compared with SPFSplat, our method yields weaker pose accuracy but substantially better novel view synthesis quality.
We attribute the pose gap to the strong MASt3R-based\cite{mast3r} initialization used by SPFSplat, which introduces additional geometric priors through correspondence-based supervision and therefore makes the comparison less directly aligned with a fully self-supervised setting.
We do not include RayZer\cite{jiang2025rayzer} and E-RayZer\cite{zhao2026erayzer} in this comparison, since pretrained weights on Re10K are not publicly available for these methods.

We next evaluate on DL3DV~\cite{ling2024dl3dv}, which provides a more challenging benchmark due to its greater scene diversity and more varied camera motion. As shown in the first two rows of \cref{tab:train_dl3dv_comparison}, compared with E-RayZer, our method achieves overall comparable pose accuracy while improving novel view synthesis quality by a clear margin. This suggests that our learned representation is more effective for rendering under self-predicted cameras. We further provide qualitative comparisons in \cref{fig:nvs_and_pose}. 
We take E-RayZer as the primary comparison baseline on this benchmark, since it is the most relevant large-scale self-supervised method and is explicitly designed to address the geometric limitations of RayZer. It is also worth noting that RayZer shows a clear dependence on sequence regularity, as its image-index embeddings provide a strong cue for shortcut learning via frame interpolation. Consistent with this observation, RayZer performs reasonably well under the sequential setting, but fails when the views are randomly ordered.

Beyond the in-domain comparison on DL3DV, we further study whether the learned representation can generalize to unseen datasets. 
The cross-dataset results in the remaining rows of \cref{tab:train_dl3dv_comparison} show that our method consistently demonstrates stronger generalization in novel view synthesis than E-RayZer, while maintaining broadly comparable pose accuracy. Notably, these evaluation image sets do not assume any fixed temporal or pose ordering, but are instead constructed to better reflect realistic multi-view captures. Since RayZer relies on ordered inputs and is therefore not suitable for this unordered cross-dataset setting, we do not treat it as a primary comparison method in the following evaluations.

\textbf{Scaling to more training data.}
We further investigate whether enlarging the training distribution with mixed-source data can improve cross-dataset generalization. 
Comparing \cref{tab:mix_comparison} with the DL3DV-only results in \cref{tab:train_dl3dv_comparison}, we find that naively training on the mixed-source set does not consistently outperform single-source training for either E-RayZer or our method. Although our method still maintains clear advantages over E-RayZer in novel view synthesis on most benchmarks, the gains brought by mixed-source training are not stable, and in several cases the performance is even weaker than that of the DL3DV-only model. These observations suggest that simply increasing training data diversity does not automatically lead to better cross-dataset transfer for pose-free reconstruction and synthesis. A likely reason is that the mixed setting introduces substantially larger variations in scene statistics, camera motion patterns, and image distributions across datasets, making it harder for the model to learn a unified representation that generalizes well across domains. Overall, while our method remains competitive under mixed-source training, robust generalization across heterogeneous data sources likely requires more dedicated mechanisms than straightforward dataset mixing alone.

\textbf{Comparison with supervised methods.}
We further compare our method with pose-supervised baselines on Re10K. 
As shown in \cref{tab:supervised_compare}, both LVSM~\cite{jin24lvsm:} and MVSplat~\cite{chen24mvsplat:} rely on camera poses during both training and inference, whereas our method is trained without pose annotations and does not require camera poses at test time.
Despite this weaker supervision, our method achieves the best PSNR and SSIM by a clear margin, outperforming both supervised baselines in reconstruction fidelity. 
Compared with LVSM, our method still achieves the best PSNR/SSIM and competitive LPIPS, indicating that it can learn an effective scene representation for high-quality novel view synthesis without explicit pose supervision. 
These results suggest that strong multi-view rendering quality can emerge from self-supervised learning alone, without relying on posed inputs or camera labels.

\begin{table}[!t]
\centering
\small
\resizebox{\columnwidth}{!}{
\begin{tabular}{llcccccc}
\toprule
\multirow{2}{*}{Dataset} & \multirow{2}{*}{Method} 
& \multicolumn{3}{c}{Pose Accuracy} 
& \multicolumn{3}{c}{Novel View Synthesis} \\
\cmidrule(lr){3-5} \cmidrule(lr){6-8}
& & @5$^\circ\uparrow$ & @15$^\circ\uparrow$ & @30$^\circ\uparrow$ & PSNR$\uparrow$ & LPIPS$\downarrow$ & SSIM$\uparrow$ \\
\midrule

\multirow{2}{*}{DL3DV\cite{ling2024dl3dv}}
& E-RayZer & 0.551 & 0.766 & 0.861 & 20.450 & 0.305 & 0.666 \\
& Ours     & 0.514 & 0.824 & 0.893 & 24.042 & 0.237 & 0.763 \\
\midrule

\multirow{2}{*}{NRGBD\cite{nrgbd_dataset_cvpr22}}
& E-RayZer & 0.280 & 0.782 & 0.913 & 25.837 & 0.173 & 0.842 \\
& Ours     & 0.360 & 0.885 & 0.984 & 30.879 & 0.114 & 0.922 \\
\midrule

\multirow{2}{*}{ScanNet++\cite{yeshwanthliu2023scannetpp}}
& E-RayZer & 0.011 & 0.262 & 0.545 & 21.950 & 0.267 & 0.743 \\
& Ours     & 0.033 & 0.284 & 0.588 & 22.191 & 0.293 & 0.749 \\
\midrule

\multirow{2}{*}{7Scenes\cite{7scenes}}
& E-RayZer & 0.323 & 0.696 & 0.966 & 26.778 & 0.184 & 0.864 \\
& Ours     & 0.296 & 0.717 & 0.791 & 28.767 & 0.157 & 0.898 \\
\bottomrule
\end{tabular}
}
\caption{Comparison under the mixed-source training setting.}
\label{tab:mix_comparison}
\vspace{-2em}
\end{table}

\begin{figure}[!t]
    \centering
    \includegraphics[width=0.99\columnwidth]{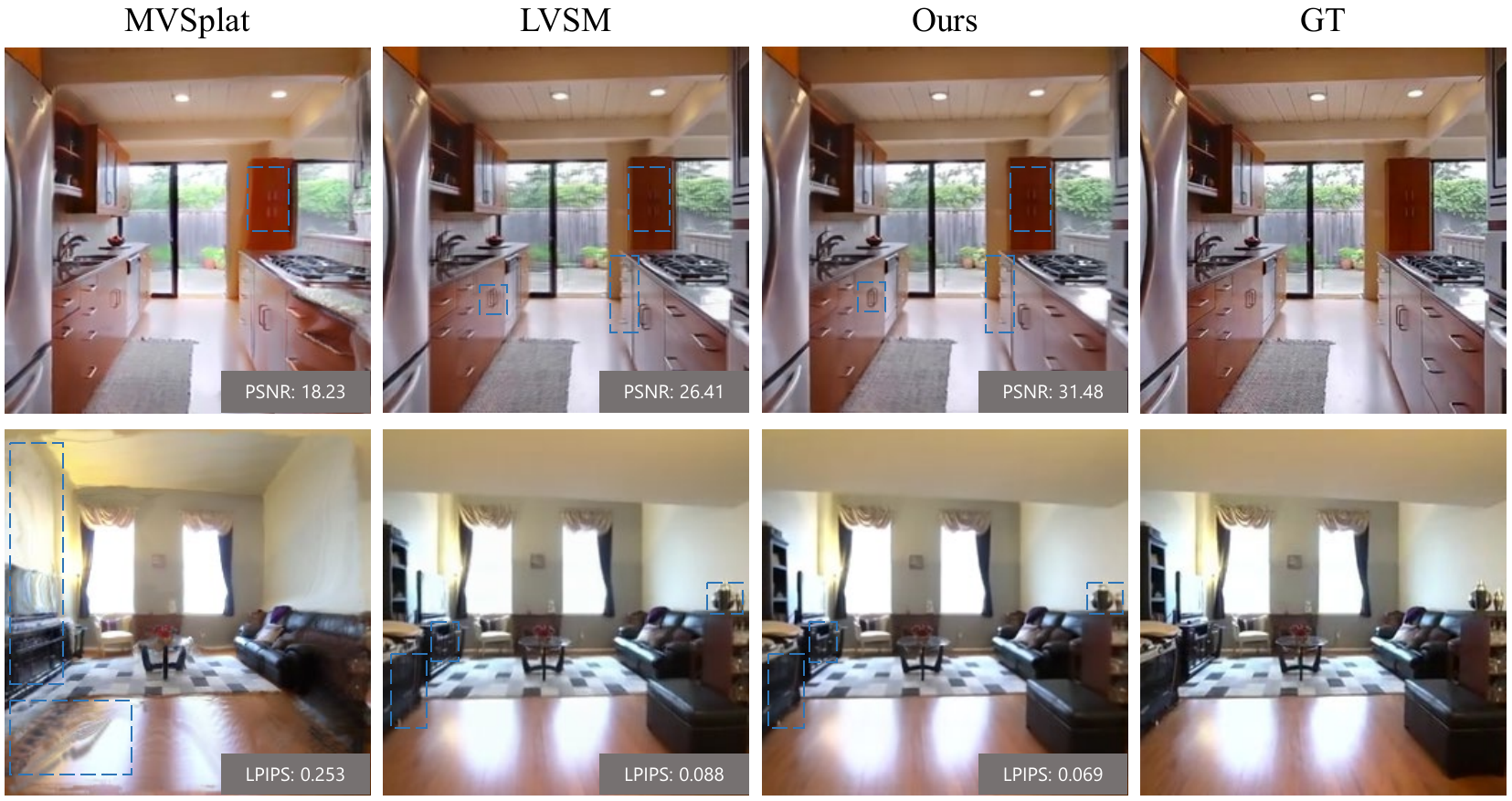}
    \caption{Qualitative comparison with supervised methods LVSM\cite{jin24lvsm:} and MVSplat\cite{chen24mvsplat:}.}
    \label{fig:compare_with_supervised}
    \vspace{-1em}
\end{figure}
\begin{table}[t]
\centering
\small
\setlength{\tabcolsep}{5pt}
\begin{tabular}{lccccc}
\toprule
\thead{Method} & 
\thead{Training\\Supervision} & 
\thead{Inference w.\\Camera Poses} & 
\thead{PSNR$\uparrow$} & 
\thead{SSIM$\uparrow$} & 
\thead{LPIPS$\downarrow$} \\
\midrule
LVSM\cite{jin24lvsm:}    & 2D + Camera & \cmarkpure & 27.603 & 0.866 & 0.110 \\
MVSplat\cite{chen24mvsplat:} & 2D + Camera & \cmarkpure & 20.257 & 0.747 & 0.241 \\
Ours    & 2D          & \xmarkpure     & 31.272 & 0.920 & 0.117 \\
\bottomrule
\end{tabular}
\caption{Comparison with pose-supervised baselines on Re10K\cite{re10k}.}
\label{tab:supervised_compare}
\vspace{-2em}
\end{table}

\subsection{Ablation studies}

\textbf{Backbone initialization.}
\cref{tab:ablation_init_weight} shows that backbone initialization is critical for both pose estimation and novel view synthesis. Initializing from Muskie\cite{li2025muskiemultiviewmaskedimage} yields substantially stronger performance than random initialization across all metrics, indicating that the gain comes from a stronger multi-view prior. 
This choice is motivated by the fact that Muskie is encouraged to discover cross-view correspondences and to produce view-consistent features with stronger geometric awareness.
Such a property is particularly suitable for our setting: the backbone is expected not only to extract per-view appearance features, but also to provide a feature space in which information from different views can be reliably aligned and aggregated.
This initialization plays an important practical role in stabilizing optimization and improving reconstruction quality.
Besides, freezing the Muskie backbone causes performance to collapse, showing that effective initialization alone is insufficient and that adapting the backbone to the target task is essential. We also find that DINOv3 performs poorly in this setting. A likely reason is that DINOv3 extracts features on a per-frame basis and lacks explicit cross-view interaction, making its features less suitable for regressing camera parameters in multi-view reconstruction.

\textbf{Renderer.}
\cref{tab:ablation_renderer} studies the effect of renderer design. Replacing our implicit renderer with a standard volume-rendering formulation still yields reasonable pose accuracy, suggesting that the model can recover useful geometric cues under this design. However, the rendering quality drops consistently across all NVS metrics, indicating that our implicit renderer provides a more effective and flexible decoding interface for high-fidelity view synthesis. We further replace the scene representation with a 3DGS-style renderer, but observe training failure even with Muskie-based initialization. This observation is consistent with our findings on RayZer\cite{jiang2025rayzer} and E-RayZer\cite{zhao2026erayzer}: although careful data scheduling can partially alleviate the issue, optimizing an explicit 3DGS representation in a fully self-supervised setting remains highly unstable.

\begin{table}[!t]
\centering
\resizebox{\columnwidth}{!}{
\small
\begin{tabular}{lccccccc}
\toprule
\multirow{2}{*}{Initialization} & \multirow{2}{*}{Frozen?}
& \multicolumn{3}{c}{Pose Accuracy}
& \multicolumn{3}{c}{Novel View Synthesis} \\
\cmidrule(lr){3-5} \cmidrule(lr){6-8}
& & @5$^\circ\uparrow$ & @15$^\circ\uparrow$ & @30$^\circ\uparrow$
& PSNR$\uparrow$ & LPIPS$\downarrow$ & SSIM$\uparrow$ \\
\midrule
Random   & \xmarkpure & 0.197 & 0.459 & 0.589 & 22.851 & 0.300 & 0.707 \\
DINOv3\cite{simeoni2025dinov3}   & \xmarkpure & 0.014    & 0.151    & 0.322    & 11.523    & 0.561    & 0.200    \\
Muskie\cite{li2025muskiemultiviewmaskedimage}   & \cmarkpure & 0.002 & 0.038 & 0.167 & 15.060 & 0.563 & 0.284 \\
Muskie\cite{li2025muskiemultiviewmaskedimage}   & \xmarkpure & 0.563 & 0.863 & 0.924 & 25.324 & 0.215 & 0.791 \\
\bottomrule
\end{tabular}
}
\caption{Ablation on backbone initialization, evaluated on DL3DV~\cite{ling2024dl3dv}.}
\label{tab:ablation_init_weight}
\vspace{-2em}
\end{table}

\begin{table}[!t]
\centering
\small
\resizebox{\columnwidth}{!}{
\begin{tabular}{lcccccc}
\toprule
\multirow{2}{*}{Method} & \multicolumn{3}{c}{Pose Accuracy} & \multicolumn{3}{c}{Novel View Synthesis} \\
\cmidrule(lr){2-4} \cmidrule(lr){5-7}
& @5$^\circ\uparrow$ & @15$^\circ\uparrow$ & @30$^\circ\uparrow$ & PSNR$\uparrow$ & LPIPS$\downarrow$ & SSIM$\uparrow$ \\
\midrule
3DGS & \multicolumn{6}{c}{\textit{Fail}} \\
Volume Rendering & 0.494 & 0.829 & 0.990 & 28.058 & 0.176 & 0.868 \\
Ours & 0.497 & 0.877 & 0.995 & 31.272 & 0.117 & 0.920 \\
\bottomrule
\end{tabular}
}
\caption{Ablation on the renderer design, evaluated on Re10K\cite{re10k}.}
\label{tab:ablation_renderer}
\vspace{-2em}
\end{table}

\begin{table}[!t]
\centering
\small
\resizebox{\columnwidth}{!}{
\begin{tabular}{@{}llcccc@{}}
\toprule
\multirow{2}{*}{Train Setting} & \multirow{2}{*}{Method} & \multirow{2}{*}{Self-supervised ?}
& \multicolumn{3}{c}{Novel View Synthesis} \\
\cmidrule(lr){4-6}
& & & PSNR$\uparrow$ & LPIPS$\downarrow$ & SSIM$\uparrow$ \\
\midrule

\multirow{6}{*}{\shortstack[l]{Re10K\\(2-view)}}
& MVSplat\cite{chen24mvsplat:}   & \xmarkpure & 22.720 & 0.168 & 0.823 \\
& LVSM\cite{jin24lvsm:}      & \xmarkpure & 27.580 & 0.120 & 0.895 \\
& NoPoSplat\cite{ye2024noposplat} & \xmarkpure & 22.865 & 0.178 & 0.770 \\
& SPFSplat\cite{huang2025spfsplat}  & \xmarkpure & 22.851 & 0.171 & 0.771 \\
& SelfSplat\cite{kang2025selfsplat} & \cmarkpure & 20.747 & 0.253 & 0.748 \\
& Ours      & \cmarkpure & 23.448 & 0.259 & 0.782 \\
\midrule

\multirow{2}{*}{\shortstack[l]{DL3DV\\(multi-view)}}
& E-RayZer\cite{zhao2026erayzer} & \cmarkpure & 14.899 & 0.287 & 0.703 \\
& Ours     & \cmarkpure & 23.708 & 0.251 & 0.794 \\
\bottomrule
\end{tabular}
}
\caption{Ablation in the sparse-view case. Pose accuracy is omitted since it is not required in this setting. Methods such as MVSplat, LVSM, NoPoSplat, and SPFSplat rely on pose supervision or external supervised initialization.}
\label{tab:sparse_view_ablation}
\vspace{-2em}
\end{table}

\subsection{Depth Visualization}
\begin{figure}[t]
    \centering
    \includegraphics[width=\columnwidth]{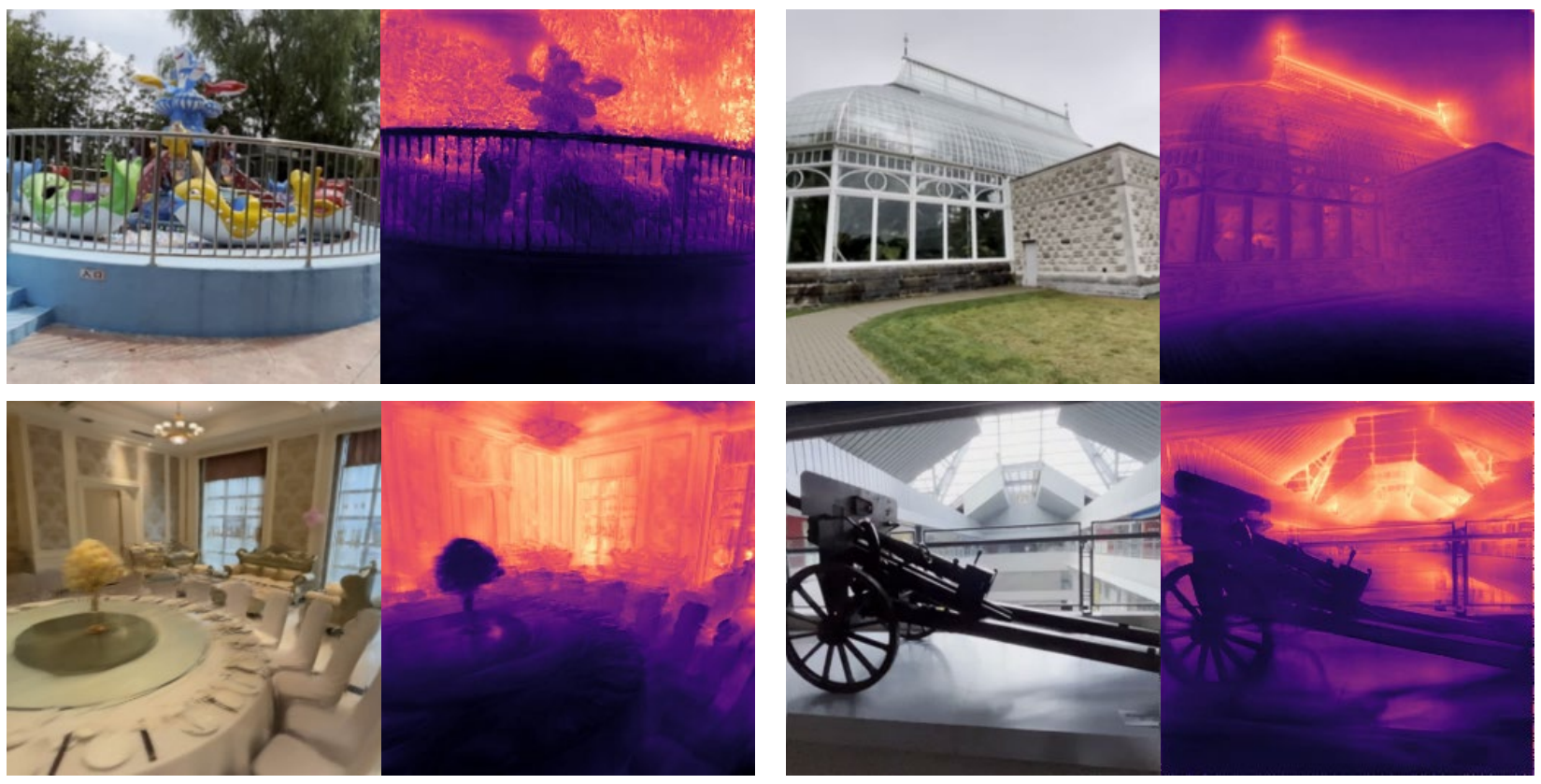}
    \caption{Rendered images and corresponding depth maps}
    \label{fig:nvs_and_depth}
\end{figure}

Although our renderer does not explicitly predict per-sample density as in classical volume rendering, we can extract a soft depth map from the ray aggregation module. For a target ray $r$, let $\{z_m\}_{m=1}^{M}$ denote the sampled depths and $\{w_m(r)\}_{m=1}^{M}$ the per-sample weights returned by the final ray-transformer block. These weights are obtained from the attention distribution of the last ray-aggregation layer and treated as a soft importance distribution. We compute
\[
D(r)=\sum_{m=1}^{M} w_m(r)z_m.
\]
Applying this computation to all target rays yields the relative depth map in Fig.~\ref{fig:nvs_and_depth}. Since training is pose-free and uses a normalized ray-sampling range, this depth should be interpreted as relative soft depth rather than metric depth.

\textbf{Sparse-view case.}
\cref{tab:sparse_view_ablation} reports results on Re10K with only two input views. 
In this highly sparse setting, methods that rely on pose supervision or external supervised initialization, such as LVSM\cite{jin24lvsm:}, MVSplat\cite{chen24mvsplat:}, NoPoSplat\cite{ye2024noposplat}, and SPFSplat\cite{huang2025spfsplat}, remain strong baselines, with LVSM achieving the best overall performance. Among the self-supervised methods trained directly in the 2-view regime, our method consistently outperforms SelfSplat\cite{kang2025selfsplat} and achieves the strongest PSNR and SSIM, although there is still a gap to the best pose-supervised model. This suggests that the extreme two-view setting remains challenging for fully self-supervised reconstruction.
More importantly, when trained with multi-view inputs and tested with only two views, our method clearly surpasses E-RayZer\cite{zhao2026erayzer} by a large margin. This indicates that the scene representation learned from multi-view training transfers more effectively to sparse-view inference, and retains strong rendering capability even when the number of available input views is drastically reduced.









\section{Conclusion}
We presented IRIS, a fully self-supervised framework for novel view synthesis from unposed multi-view images. IRIS represents the scene as a latent neural field queried and rendered under self-predicted cameras, providing a practical middle ground between flexible latent modeling and geometrically grounded 3D reasoning. Experiments demonstrate strong rendering quality, competitive pose accuracy, and favorable generalization. We hope this work offers useful insights to the community on designing future self-supervised 3D vision models.

\section{Appendix}









\begin{table*}[t]
\centering
\small
\setlength{\tabcolsep}{6pt}
\renewcommand{\arraystretch}{1.28}

\begin{tabularx}{0.7\textwidth}{
>{\centering\arraybackslash}m{1.8cm}
>{\raggedright\arraybackslash}X
>{\raggedright\arraybackslash}X}
\toprule
\textbf{Method} & \textbf{Scene modeling} & \textbf{Rendering} \\
\midrule

\rowcolor{rayzergreen!6}
\textcolor{rayzergreen}{\textbf{RayZer}}
&
\makecell[l]{
$\displaystyle z_{\mathrm{scene}}
=
f_{\psi}^{\mathrm{scene}}\!\left(
z_{0}^{\mathrm{scene}},\, x_{\mathrm{ref}}
\right)$
}
&
\makecell[l]{
$\displaystyle
\hat{I}_{\mathrm{tgt}}
=
f_{\phi}^{\mathrm{rend}}\!\left(
z_{\mathrm{scene}},\,
\mathrm{Linear}\!\left(R_{\mathrm{tgt}}^{\mathrm{plk}}\right)
\right)$
}
\\

\midrule

\rowcolor{erayzerblue!6}
\textcolor{erayzerblue}{\textbf{E-RayZer}}
&
\makecell[l]{
$\displaystyle
\mathcal{G}_{\mathrm{ref}}
=
f_{\omega}^{\mathrm{gauss}}\!  \circ 
f_{\psi'}^{\mathrm{scene}}(x_{\mathrm{ref}})
$
}
&
\makecell[l]{
$\displaystyle
\hat{I}_{\mathrm{tgt}}
=
\pi\!\left(
\mathcal{G}_{\mathrm{ref}},\,
P_{\mathrm{tgt}}
\right)$
}
\\

\midrule

\rowcolor{irisorange!8}
\textcolor{irisorange}{\textbf{IRIS}}
&
\makecell[l]{
$\displaystyle
\tilde f(x)
=
\mathcal T_{\mathrm{view}}
\!\left(
v_{1}(x),\ldots,v_{V_{\mathrm{ref}}}(x)
\right)$
}
&
\makecell[l]{
$\displaystyle
\hat C(r)=f_{\mathrm{rgb}}  \circ  \mathcal T_{\mathrm{ray}}
\!\left(
\tilde f_{1},\ldots,\tilde f_{M}
\right)$
}
\\

\bottomrule
\end{tabularx}

\caption{Comparison of RayZer, E-RayZer, and IRIS in scene modeling and rendering.}
\label{tab:formal_compare_rendering}
\end{table*}
\paragraph{Definition of Pose Error.}
Given a set of predicted and ground-truth camera-to-world poses,
\(\{\hat{\mathbf T}^{\mathrm{c2w}}_i\}_{i=1}^{B}\) and
\(\{\mathbf T^{\mathrm{c2w}}_i\}_{i=1}^{B}\),
we first convert them to world-to-camera poses:
\[
\hat{\mathbf T}^{\mathrm{w2c}}_i = (\hat{\mathbf T}^{\mathrm{c2w}}_i)^{-1},
\qquad
\mathbf T^{\mathrm{w2c}}_i = (\mathbf T^{\mathrm{c2w}}_i)^{-1}.
\]
We then evaluate all unordered view pairs
\[
\mathcal P=\{(i,j)\mid 1\le i<j\le B\}.
\]
For each pair \((i,j)\in\mathcal P\), the relative pose is defined as
\[
\hat{\mathbf T}_{i\leftarrow j}
=
\hat{\mathbf T}^{\mathrm{w2c}}_i
\left(\hat{\mathbf T}^{\mathrm{w2c}}_j\right)^{-1},
\qquad
\mathbf T_{i\leftarrow j}
=
\mathbf T^{\mathrm{w2c}}_i
\left(\mathbf T^{\mathrm{w2c}}_j\right)^{-1}.
\]
Let
\[
\hat{\mathbf T}_{i\leftarrow j}
=
\begin{bmatrix}
\hat{\mathbf R}_{i\leftarrow j} & \hat{\mathbf t}_{i\leftarrow j}\\
\mathbf 0^\top & 1
\end{bmatrix},
\qquad
\mathbf T_{i\leftarrow j}
=
\begin{bmatrix}
\mathbf R_{i\leftarrow j} & \mathbf t_{i\leftarrow j}\\
\mathbf 0^\top & 1
\end{bmatrix},
\]
where \(\hat{\mathbf R}_{i\leftarrow j},\mathbf R_{i\leftarrow j}\in SO(3)\)
and \(\hat{\mathbf t}_{i\leftarrow j},\mathbf t_{i\leftarrow j}\in\mathbb R^3\).

The rotation error is measured by the geodesic angle between the predicted and ground-truth relative rotations:
\[
e_R(i,j)
=
\arccos\!\left(
\frac{
\operatorname{Tr}\!\left(\mathbf R_{i\leftarrow j}^{\top}\hat{\mathbf R}_{i\leftarrow j}\right)-1
}{2}
\right)\cdot \frac{180}{\pi}.
\]

For translation, we only evaluate the direction and ignore the global scale. Let
\[
c_t(i,j)=
\frac{
\mathbf t_{i\leftarrow j}^{\top}\hat{\mathbf t}_{i\leftarrow j}
}{
\|\mathbf t_{i\leftarrow j}\|_2\,
\|\hat{\mathbf t}_{i\leftarrow j}\|_2
}.
\]
Since the relative translation is treated as sign-ambiguous in our implementation, the translation-direction error is defined as
\[
e_t(i,j)
=
\min\!\bigl(\arccos(c_t(i,j)),\, \pi-\arccos(c_t(i,j))\bigr)\cdot\frac{180}{\pi}.
\]
Equivalently, this can be viewed as the unsigned angular error between the two translation directions.

Following the implementation, the \textbf{relative pose error} for each pair is defined as the maximum of the rotation and translation-direction errors:
\[
e_{\mathrm{pose}}(i,j)=\max\!\bigl(e_R(i,j),\,e_t(i,j)\bigr).
\]

Given an angular threshold \(\delta\in\{5^\circ,15^\circ,30^\circ\}\), the relative pose accuracy (RPA) is computed as
\[
\mathrm{RPA}@\delta
=
\frac{1}{|\mathcal P|}
\sum_{(i,j)\in\mathcal P}
\mathbf 1\!\left[e_{\mathrm{pose}}(i,j)\le \delta\right],
\]
where \(\mathbf 1[\cdot]\) denotes the indicator function.

\paragraph{Compare with standard volume rendering.}

Both standard volume rendering and our method can be written under a unified ray-decoding form:
\[
\hat C(r)=\mathcal R\bigl(\{g(x_m,r)\}_{m=1}^{M}\bigr),
\qquad
x_m=o_r+z_m d_r,
\]
where \(r=(o_r,d_r)\) is a target ray, \(\{x_m\}_{m=1}^{M}\) are the sampled 3D points along the ray, \(g(x_m,r)\) denotes the per-sample representation to be composed, and \(\mathcal R(\cdot)\) denotes the ray-wise composition rule.

For standard volume rendering, the scene is parameterized by explicit radiance-field quantities:
\[
g_{\mathrm{vol}}(x_m,r)=(\sigma_m,c_m),
\qquad
(\sigma_m,c_m)=F_{\mathrm{vol}}(x_m,d_r),
\]
where \(\sigma_m\) and \(c_m\) denote the density and color at the \(m\)-th sample. The final pixel color is obtained by a fixed hand-crafted compositing rule:
\[
\hat C_{\mathrm{vol}}(r)
=
\sum_{m=1}^{M} T_m \alpha_m c_m,
\]
with
\[
\alpha_m = 1-\exp(-\sigma_m \delta_m),
\qquad
T_m=\prod_{n=1}^{m-1}(1-\alpha_n).
\]

In contrast, our method does not predict explicit density and color at each sampled point. Instead, it constructs a latent field from multi-view features and uses the queried latent feature as the per-sample representation:
\[
g_{\mathrm{IRIS}}(x_m,r)=\tilde f_m,
\qquad
\tilde f_m = F_{\mathcal S}(x_m),
\]
where
\[
F_{\mathcal S}(x_m)
=
\mathcal T_{\mathrm{view}}\!\bigl(v_1(x_m),\dots,v_{V_{\mathrm{ref}}}(x_m)\bigr).
\]
These point-wise latent features are then composed along the ray by a learnable ray-wise aggregator:
\[
h_r=\mathcal T_{\mathrm{ray}}(\tilde f_1,\dots,\tilde f_M),
\qquad
\hat C_{\mathrm{IRIS}}(r)=f_{\mathrm{rgb}}(h_r).
\]

Therefore, the key difference is that standard volume rendering uses explicit physical quantities \((\sigma,c)\) together with a fixed alpha-compositing rule, whereas our method uses latent point-wise features together with a learned ray-wise composition function.

\begin{figure}[!t]
    \centering
    \includegraphics[width=0.99\columnwidth]{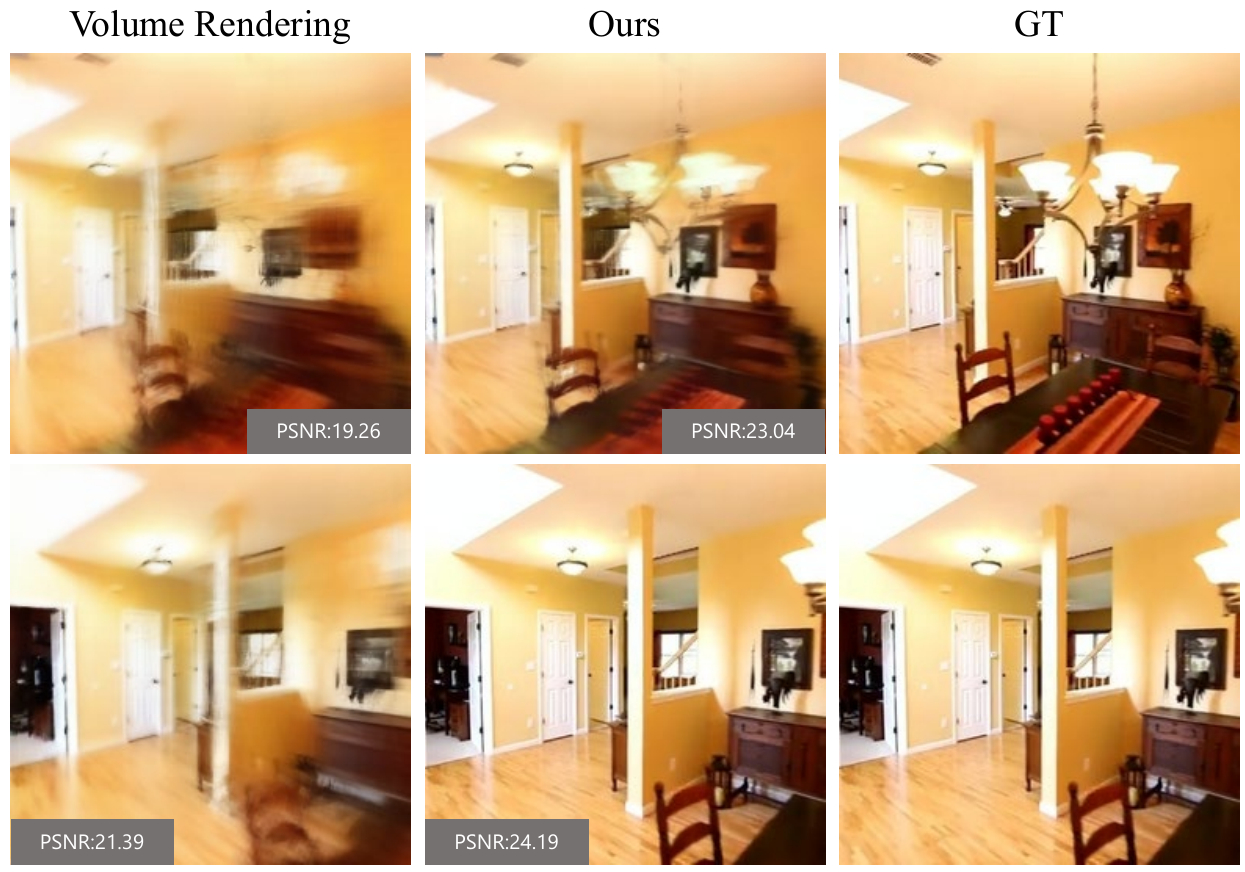}
    \caption{Compared with standard volume rendering, implicit rendering improves quality of novel view synthesis.}
    \label{fig:compare_renderer}
    \vspace{-1em}
\end{figure}

Besides Table 7 in the main paper, \cref{fig:compare_renderer} shows that replacing our learned implicit renderer with standard volume rendering produces visibly blurrier results and consistently worse quantitative performance. We attribute this gap to the fact that volume rendering uses a fixed hand-crafted composition rule, while our renderer learns to compose view-conditioned latent features adaptively along each ray, yielding a more expressive decoding interface for high-fidelity synthesis.
This phenomenon has also been observed in prior work. For example, GNT\cite{GNT,gntmove2023} showed that learned ray-wise rendering can better capture fine structures and appearance effects. 
Our result extends this observation to a new regime: unlike prior evidence obtained in posed settings, we verify it in a fully pose-free setting, where camera parameters themselves are learned under self-supervision. 

\paragraph{Formal comparison of scene modeling and rendering}
Table~\ref{tab:formal_compare_rendering} summarizes the main differences among RayZer, E-RayZer, and IRIS in scene modeling and rendering. RayZer represents the scene with latent tokens and renders target views using a learned decoder. E-RayZer instead predicts explicit 3D Gaussians and renders them by differentiable splatting. In contrast, IRIS induces a latent neural field from multi-view features and performs learnable ray-wise rendering. This comparison highlights that our method differs from prior self-supervised paradigms in both the scene representation and the rendering interface.

\begin{figure*}[t]
    \centering
    \includegraphics[width=0.8\textwidth]{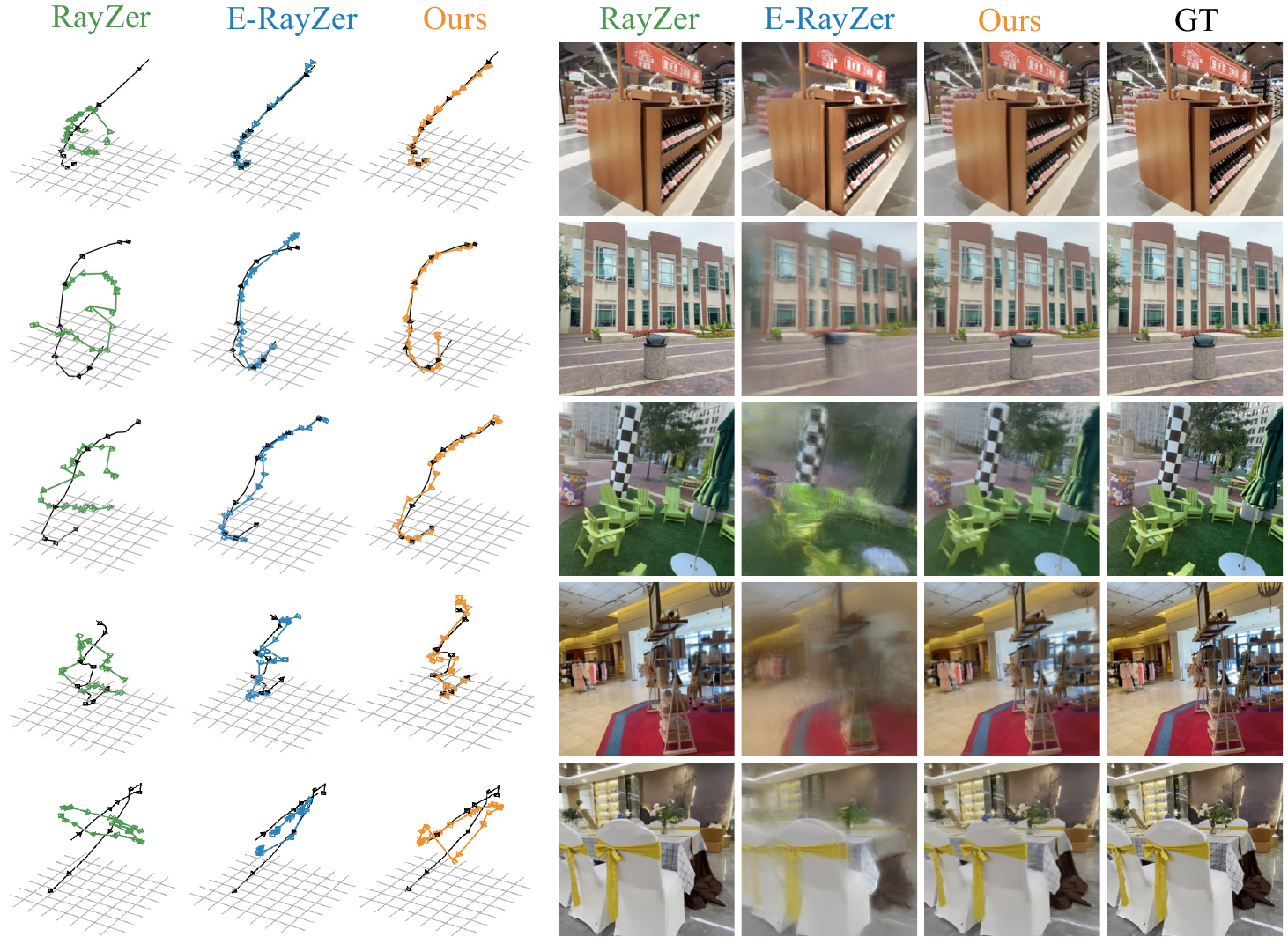}
    \caption{More qualitative comparisons of pose estimation and rendering quality. We visualize all poses of the predicted trajectory as colored frustums, while for the COLMAP trajectory we display only every third pose as a black frustum to reduce overlap.}
    \label{fig:nvs_and_pose_supply}
\end{figure*}

\paragraph{More Qualitative comparisons}
We present more qualitative comparisons in \cref{fig:nvs_and_pose_supply}.
x

\begin{algorithm}[!t]
\caption{Computing point-wise latent features}
\label{alg:view_aggregation_expanded}
\begin{algorithmic}[1]
\Require A 3D query point $x$; latent scene memory
\[
\mathcal{S}=\{(f_k,K,P_k)\}_{k\in\mathcal{V}_{\mathrm{ref}}}
\]
\Require View-transformer blocks $\{\mathcal T_{\mathrm{view}}^{(l)}\}_{l=1}^{L}$

\For{each reference view $k\in\mathcal{V}_{\mathrm{ref}}$}
    \State Project $x$ into view $k$ and sample the corresponding feature:
    \[
    v_k(x)=\Pi_k(x;f_k,P_k,K)
    \]
    \State Compute the relative geometric cue $\Delta_k(x)$ between the target ray and view $k$
    \State Compute the visibility / validity mask $m_k(x)$
\EndFor

\State Project sampled view-wise features to the model width:
\[
u_k^{(0)}(x)=\phi_{\mathrm{proj}}\!\bigl(v_k(x)\bigr), \qquad k=1,\dots,V_{\mathrm{ref}}
\]
\State Initialize by max-pooling across reference views:
\[
q^{(0)}(x)=\max_{k\in\mathcal{V}_{\mathrm{ref}}} u_k^{(0)}(x)
\]

\For{$l=1$ to $L$}
    \State Update the point token by aggregating multi-view features:
    \[
    q^{(l)}(x)
    =
    \mathcal T_{\mathrm{view}}^{(l)}
    \bigl(
    q^{(l-1)}(x),\,
    \{u_k^{(0)}(x)\}_{k=1}^{V_{\mathrm{ref}}},\,
    \{\Delta_k(x)\}_{k=1}^{V_{\mathrm{ref}}},\,
    \{m_k(x)\}_{k=1}^{V_{\mathrm{ref}}}
    \bigr)
    \]
\EndFor

\State Define the point-wise latent feature as
\[
\tilde f(x)=F_{\mathcal S}(x)=q^{(L)}(x)
\]
\State \Return $\tilde f(x)$
\end{algorithmic}
\end{algorithm}

\begin{algorithm}[!t]
\caption{Rendering a target ray from point-wise latent features}
\label{alg:ray_aggregation_expanded}
\begin{algorithmic}[1]
\Require A target ray $r=(o_r,d_r)$
\Require Precomputed point-wise latent features $\{\tilde f_m^{(0)}\}_{m=1}^{M}$ and their sample locations $\{x_m\}_{m=1}^{M}$
\Require Ray-transformer blocks $\{(\psi^{(l)},\,\mathcal T_{\mathrm{ray}}^{(l)})\}_{l=1}^{L}$

\State Encode the sample locations and the target-ray direction:
\[
e_m^{\mathrm{pos}}=\gamma_{\mathrm{pos}}(x_m),\qquad
e^{\mathrm{dir}}=\gamma_{\mathrm{dir}}(d_r)
\]

\For{$l=1$ to $L$}
    \State Update each point feature with point/ray encodings:
    \[
    \tilde f_m^{(l-\frac12)}
    =
    \psi^{(l)}\!\bigl(\tilde f_m^{(l-1)},\,e_m^{\mathrm{pos}},\,e^{\mathrm{dir}}\bigr),
    \qquad m=1,\dots,M
    \]
    \State Aggregate the point-wise features along the ray:
    \[
    (\tilde f_1^{(l)},\dots,\tilde f_M^{(l)})
    =
    \mathcal T_{\mathrm{ray}}^{(l)}
    \bigl(
    \tilde f_1^{(l-\frac12)},\dots,\tilde f_M^{(l-\frac12)}
    \bigr)
    \]
\EndFor

\State Normalize and pool the final point features:
\[
h_r
=
\frac{1}{M}\sum_{m=1}^{M}
\mathrm{Norm}\!\bigl(\tilde f_m^{(L)}\bigr)
\]
\State Predict the final ray color:
\[
\hat C(r)=f_{\mathrm{rgb}}(h_r)
\]
\State \Return $\hat C(r)$
\end{algorithmic}
\end{algorithm}

\paragraph{Implementation details.}
For ray sampling, we use a fixed normalized depth range of $[0.1, 1.0]$ in all experiments. We note that, in the fully self-supervised pose-free setting, the recovered camera translations are only defined up to a global scale, so the absolute scene depth is not directly meaningful. In this sense, the depth range mainly serves as a canonical sampling interval rather than a metric depth prior. We choose $[0.1, 1.0]$ empirically, and find it sufficient for stable training and rendering quality across datasets.
Our implementation is mainly based on the GNT-style aggregation design. However, unlike GNT\cite{GNT}, which alternates multiple view and ray transformer blocks, our renderer uses only a single view aggregation stage to compute point-wise latent features and a single ray aggregation stage to decode the final ray color.
In practice, the point-wise latent feature computation starts by projecting a 3D query point $x$ into each reference view and sampling the corresponding feature
\[
v_k(x)=\Pi_k(x;f_k,P_k,K), \qquad k\in\mathcal V_{\mathrm{ref}}.
\]
Besides the sampled feature itself, we also compute a relative geometric cue $\Delta_k(x)$ and a binary validity mask $m_k(x)$. The cue $\Delta_k(x)$ provides geometry-aware information about the relation between the target ray and the $k$-th reference-view observation, while $m_k(x)$ indicates whether the projected point falls into a valid image region and should participate in aggregation. The sampled features are first projected to the model width and max-pooled across views for initialization, and are then fused by a single view transformer to produce the point-wise latent feature $\tilde f(x)=F_{\mathcal S}(x)$.

For ray aggregation, given sampled points $\{x_m\}_{m=1}^{M}$ on a target ray $r=(o_r,d_r)$, we first compute their point-wise latent features $\{\tilde f_m\}_{m=1}^{M}$ and then inject sinusoidal encodings of both the sample locations and the target-ray direction:
\[
e_m^{\mathrm{pos}}=\gamma_{\mathrm{pos}}(x_m),\qquad
e^{\mathrm{dir}}=\gamma_{\mathrm{dir}}(d_r).
\]
These enriched point features are then composed by a single ray transformer into a ray-wise representation $h_r$, from which the final color is predicted by $f_{\mathrm{rgb}}$.

\begin{table}[!t]
\centering
\small
\setlength{\tabcolsep}{4pt}
\renewcommand{\arraystretch}{1.12}
\resizebox{\columnwidth}{!}{
\begin{tabular}{p{1.65cm} c c p{2.8cm}}
\toprule
\textbf{Method} & \textbf{Self-supervised?} & \makecell{\textbf{Mutli-view}\\\textbf{capable?}} & \textbf{Remarks} \\
\midrule

MVSplat   & \xmark & \xmark & \\
LVSM      & \xmark & \cmark & \\
\midrule

PF-LRM    & \xmark & \cmark & Not publicly available. \\
NoPoSplat & \xmark & \xmark & \\
\midrule

SelfSplat & \cmark & \xmark & \\
SPFSplat  & \xmark(MASt3R init) & \cmark & Only trained on Re10K \\ 
RayZer    & \cmark & \cmark & \\
E-RayZer  & \cmark & \cmark & \\
\bottomrule
\end{tabular}
}
\caption{Taxonomy of the main baselines discussed in this paper, following our related-work categorization.}
\label{tab:baseline_taxonomy}
\end{table}

\begin{table}[!t]
\centering
\footnotesize
\resizebox{0.8\columnwidth}{!}{
\begin{tabular}{ccrrr}
\toprule
$M$ & $N$ & E-RayZer (ms) & RayZer (ms) & IRIS (s)\\
\midrule
4 & 1 & 31.27 & 36.37 & 2.74  \\
8 & 2 & 32.77 & 52.01 & 7.92  \\
12 & 4 & 44.67 & 80.63 & 20.75  \\
16 & 8 & 70.96 & 133.35 & 49.24  \\
\bottomrule
\end{tabular}
}
\caption{Inference latency for baselines and ours.}
\label{tab:efficiency_compact}
\end{table}

\paragraph{Inference efficiency.}
To evaluate inference efficiency, we benchmark all methods on a fixed test scene and vary the numbers of context and target views, denoted by $M$ and $N$, respectively. For each $(M,N)$ setting, we construct a single-scene batch by selecting $M$ context frames and $N$ target frames, move the batch to GPU, and measure the forward-pass latency using CUDA events. Following the benchmark script, each configuration is first warmed up for two runs and then timed for four runs, and we report representative mean latency values in Table~\ref{tab:efficiency_compact}. By default, the benchmark measures the model forward pass itself and excludes additional method-specific input preparation overhead, so the reported numbers primarily reflect the rendering cost of each method under the corresponding $(M,N)$ configuration.
As shown in Table~\ref{tab:efficiency_compact}, E-RayZer is the fastest among the three methods at all representative operating points, while RayZer is moderately slower but remains in the millisecond regime. Our method is noticeably more expensive, and its latency increases more rapidly as both $M$ and $N$ grow. This trend is expected, since our renderer explicitly performs learned view aggregation and ray aggregation. As a result, the computational cost grows with both the number of target views and the amount of reference views used for each ray.
At the same time, we stress that this additional cost is closely tied to the stronger rendering capability of our method. Our method adopts a finer-grained rendering process over sampled 3D points and multi-view features. In this sense, the higher latency mainly reflects a quality--efficiency trade-off.

\bibliographystyle{ACM-Reference-Format}
\bibliography{references,refs_my,refs,vedaldi_general,vedaldi_specific}
\end{document}